\pdfoutput=1

\documentclass[11pt]{article}

\usepackage[]{acl}

\usepackage{times}
\usepackage{latexsym}
\usepackage{algorithm}
\usepackage{algpseudocode}

\usepackage[T1]{fontenc}

\usepackage[utf8]{inputenc}
\usepackage{amsmath} 

\usepackage{microtype}

\usepackage{inconsolata}

\usepackage{graphicx}
\usepackage{booktabs}
\usepackage{tabularx}
\usepackage{multirow}
\usepackage{array}
\usepackage{bbding}
\usepackage{changes}
\usepackage{amsmath, amssymb}
\usepackage[T1]{fontenc}
\usepackage[utf8]{inputenc}
\usepackage{authblk}
\usepackage{hyperref}

\title{VisLingInstruct: Elevating Zero-Shot Learning in Multi-Modal Language Models with Autonomous Instruction Optimization}

\author{
    Dongsheng Zhu \textsuperscript{\rm1},
    Xunzhu Tang \textsuperscript{\rm2},
    Weidong Han \textsuperscript{\rm3}, 
    Jinghui Lu \textsuperscript{\rm4}, 
    \\
    Yukun Zhao \textsuperscript{\rm1},
    Guoliang Xing \textsuperscript{\rm1}, 
    Junfeng Wang \textsuperscript{\rm1}, 
    Dawei Yin$^*$ \textsuperscript{\rm1} 
    \\
    \textsuperscript{1} Baidu Inc., \textsuperscript{2} University of Luxemburg \\
    \textsuperscript{3} Fudan University, \textsuperscript{4} University College Dublin \\
    \texttt{\{zhudongsheng, yindawei02\}@baidu.com} \\
    \texttt{xunzhu.tang@uni.lu, wdhan20@fudan.edu.cn}
}

\begin{document}
\maketitle

\def\thefootnote{*}\footnotetext{Corresponding author}

\begin{abstract}
 This paper presents VisLingInstruct, a novel approach to advancing Multi-Modal Language Models (MMLMs) in zero-shot learning. Current MMLMs show impressive zero-shot abilities in multi-modal tasks, but their performance depends heavily on the quality of instructions. VisLingInstruct tackles this by autonomously evaluating and optimizing instructional texts through In-Context Learning, improving the synergy between visual perception and linguistic expression in MMLMs. Alongside this instructional advancement, we have also optimized the visual feature extraction modules in MMLMs, further augmenting their responsiveness to textual content. Our comprehensive experiments on MMLMs, based on FlanT5 and Vicuna, show that VisLingInstruct significantly improves zero-shot performance in visual multi-modal tasks. Notably, it achieves a 13.1\% and 9\% increase in accuracy over the prior state-of-the-art on the TextVQA and HatefulMemes datasets. Our main code  is available at \url{https://github.com/Zhudongsheng75/VisLingInstruct}.
\end{abstract}
\section{Introduction}

The integration of Large Language Models (LLMs) with vision or other multi-modalities, epitomized by models like BLIP-2 \citep{chen2022pali,alayrac2022flamingo,li2023blip}, has marked a significant evolution in the Natural Language Processing (NLP) field. This advancement led to the emergence of Multi-Modal Language Models (MMLMs), blending visual and linguistic data processing to enhance complex multimodal information understanding and generation. InstructBLIP \cite{instructblip}, a notable example, utilizes advanced instruction tuning for image-text pairs, significantly improving the Q-Former module's zero-shot learning capabilities in a variety of vision-language tasks. This progression underscores the potential of MMLMs in navigating the intricacies of multi-modal data, setting a new benchmark in the intersection of language, vision, and machine learning.

However, the effectiveness of MMLMs is highly constrained by the quality of textual instructions. Current instruction-tuned models \citep{ouyang2022training,zheng2023judging} are effective, while they may introduce significant challenges, particularly for users who lack expertise in crafting optimal instructions. This limitation leads to inconsistent or sub-optimal outputs, thus impeding the practical utility of MMLMs in the real world scenarios. To mitigate this issue, we propose a novel autonomous optimization method for textual instruction, named \textbf{Vis}ual, \textbf{Ling}uistic, \textbf{Instruct}ion optimization (VisLingInstruct). The VisLingInstruct introduces an innovative method via In-Context Learning (ICL) \cite{min2022rethinking} based on the comparison between instruction cases. We incorporate it with our newly proposed Instruction Alignment Score (IAS) to exploit the inherent capacity of MMLMs to self-evaluate the quality of text instructions. Consequently, VisLingInstruct can guide the model towards the generation of more effective and contextually appropriate instructions.

Complementing our instructional optimization strategy, we present an architectural innovation aimed at enhancing the alignment between visual and textual modules within MMLMs. Inspired by recent advancements in models such as Mini-GPT4 \citep{zhu2023minigpt}, LLaVA \citep{liu2023visual}, mPLUG-Owl \citep{ye2023mplug}, and BLIVA \cite{hu2023bliva}, our architecture enhances the integration of textual and visual data. Our new approach enables MMLMs to more effectively process complex tasks that require an understanding of both textual and visual elements, thereby improving accuracy and contextual understanding. Figure \ref{comparison.png} offers a visual comparison of the alignment modules in different MMLMs, highlighting the distinctive features and benefits of our proposed method. Through this architectural enhancement, we aim to bridge the existing gaps in multi-modal data processing, creating a more cohesive and efficient model capable of tackling the nuanced demands of multi-modal interactions.

In summary, our contributions are as follows:
\begin{itemize}
    \item We introduce substantial architectural improvements for better integration of multi-modal data within MMLMs for training and inference (Section~\ref{sec:architecture}). 
    \item We present an autonomous method for optimizing instruction quality, tailored to improve the effectiveness of textual instruction during inference (Section~\ref{sec:self_optimization}). To the best of our knowledge, we spearhead the manual-free optimization of textual instruction in zero-shot for multi-modal tasks.
    \item We conduct comprehensive experiments and ablation studies to demonstrate the effectiveness of VisLingInstruct and the success of each component. Notably, VisLingInstruct has improved the performance by a significant margin of 13.1\% and 9\% on the TextVQA and HatefulMemes dataset.
\end{itemize}

\section{Related Work}
\subsection{Instruction Tuning in MMLMs}
Instruction tuning has emerged as a cost-effective alternative to the expensive pre-training of large models, focusing on fine-tuning a few foundational models for downstream tasks. In this context, models like InstructGPT \citep{ouyang2022training}, Flan-T5 \citep{chung2022scaling}, and Vicuna \citep{zheng2023judging} represent significant strides in conversational models obtained through instruction tuning based on LLMs. These models have showcased exceptional question-answering capabilities, underscoring the importance of instruction-based approaches in language generation. In the multi-modal domain, advancements such as Mini-GPT4 \citep{zhu2023minigpt}, LLaVA \citep{liu2023visual}, mPLUG-Owl \citep{ye2023mplug}, InstructBLIP \citep{instructblip}, and BLIVA \cite{hu2023bliva} have focused on instruction fine-tuning. These methods typically involve aligning images and text by introducing transitional layers, like Q-Former and fully connected layers, between visual encoders and LLMs. Our work builds upon these foundations, aiming to further optimize the instruction tuning process for enhanced performance in MMLMs.

\begin{figure}[t]
\centering
\includegraphics[width=0.5\textwidth]{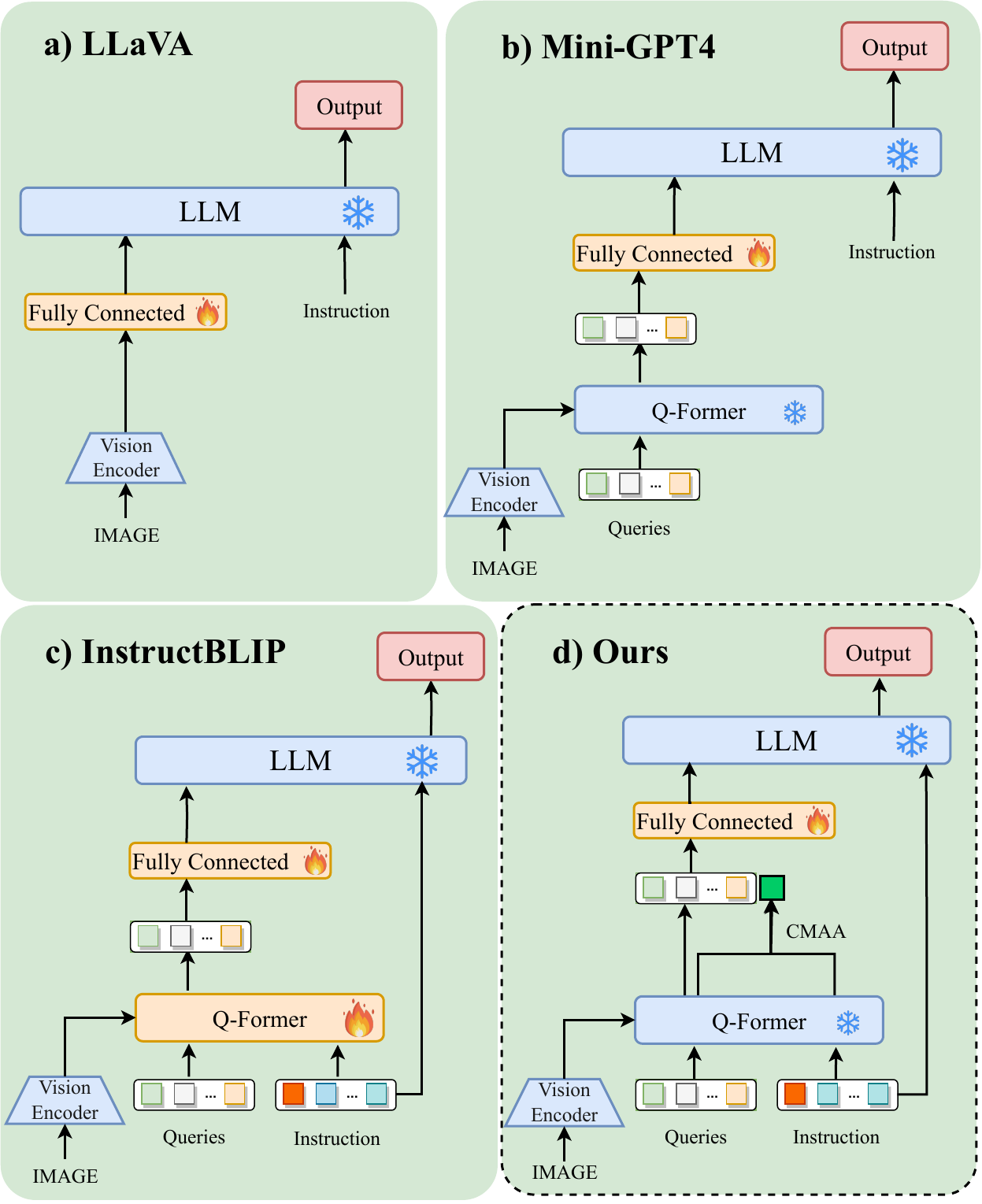}
\caption{The structural comparison among the alignment modules of different MMLMs. The orange modules in the figure represent open weights, while the blue modules indicate frozen weights.}
\label{comparison.png}
\end{figure}

\begin{figure*}[htp]
\centering
\includegraphics[width=1\textwidth]{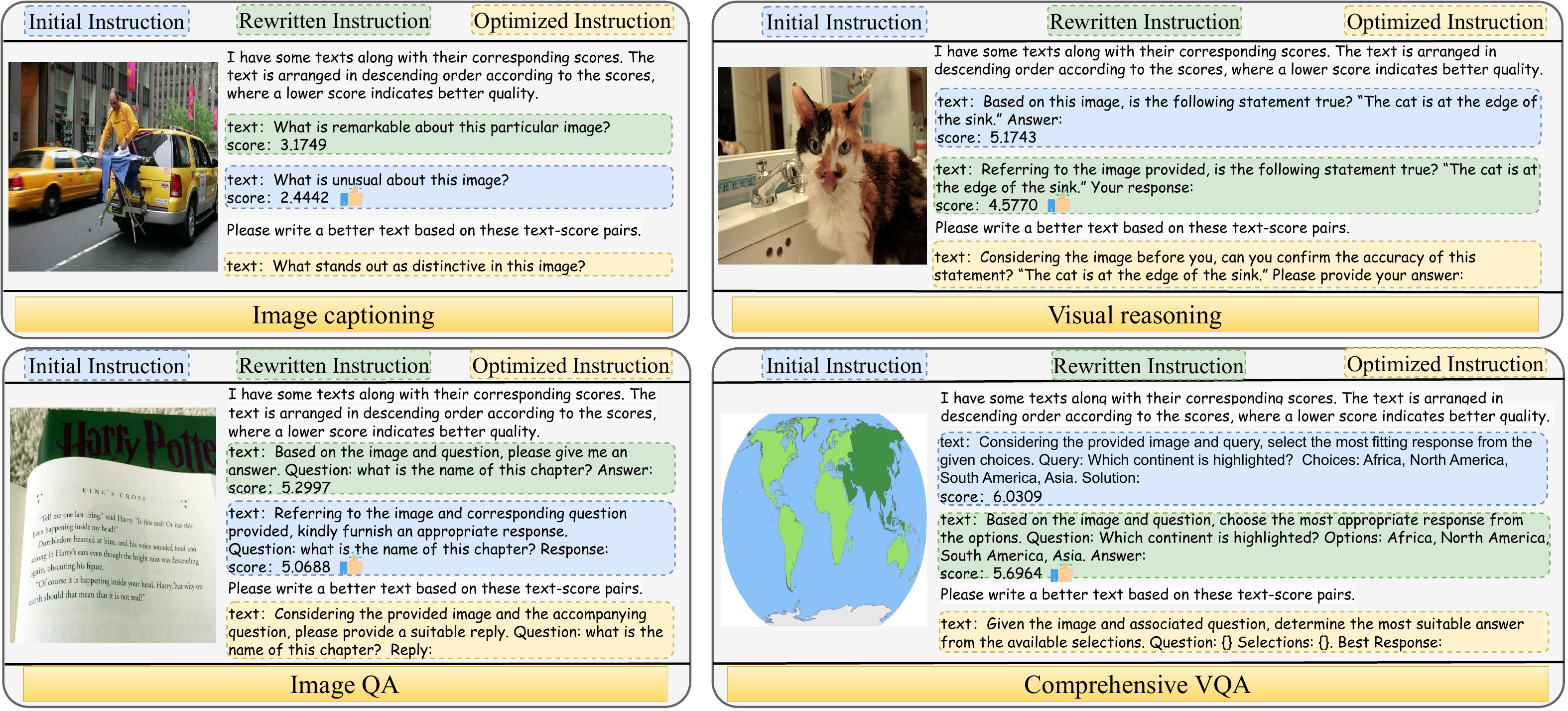}
\caption{The examples of ranking with ICL in different domains. On the left side is the image input provided to the MMLM. On the right side, within the blue box, lies the initial instruction, while the rewritten instruction is contained within the green box. The `score', referred to as IAS, indicates the quality of corresponding instructions with respect to the model, while the lower score (i.e., high quality) instruction ranks lower in the ICL demonstration. By utilizing the paradigm of ICL, MMLM learn the relationship between the scores of the two cases to generate higher-quality new instructions that lie in the yellow box.}
\label{icl.png}
\end{figure*}

\subsection{Optimizing Instructions for Large Models}
Historically, Pre-trained Language Models (PLM) akin to BERT \citep{kenton2019bert} have utilized prompt crafting techniques \cite{brown2020language,jiang2022promptbert} to boost performance, with subsequent research exploring methods to discover higher-quality prompts \cite{gao2021making,lu2023makes}. In generative models, this concept has evolved into optimizing `instructions', leading to a series of works focused on prompt and instruction optimization \cite{wei2022chain,min2022rethinking}. Notably, UPRISE \cite{cheng2023uprise} trained a prompt retriever for acquiring superior instructions, while OPRO \cite{yang2023large} conceptualized LLMs as optimizers, formulating optimization tasks in textual instructions. \cite{zheng2023take} introduced STEP-BACK prompting, enabling LLMs to derive higher-level concepts from detailed instances. 

\section{Methods}
Our approach comprises two components: First, we refine the architecture of existing multi-modal models and their fine-tuning mechanisms to augment their perceptivity of instruction, that is, the Enhanced Multi-modal Alignment (EMA). Second, subsequent to the model’s fine-tuning, we concentrate on the autonomous optimization of instructions during the inference, referred to as the Autonomous Instruction Optimization (AIO).

\subsection{Enhancing Multi-modal Alignment}
\label{sec:architecture}
In the quest to refine MMLM, our focus shifts to bridging the gap between the realms of visual perception and linguistic expression. This section delves into our approach to enhancing the alignment between visual and textual modules within MMLM, introducing the architectural innovation and training optimization designed to synergize these two distinct modalities seamlessly.

\textbf{Integrative Processing of Text and Image:}
At the core of our architectural enhancements is the integrative processing of textual and visual data. The process involves constructing a unified representation by merging detailed textual embeddings with rich visual information. We introduce the Cross-Modal Alignment Attention (CMAA) algorithm to achieve this integration, specifically designed to harmonize these disparate data modalities. CMAA leverages attention mechanisms \cite{bahdanau2014neural} and cross-modal feature fusion \cite{radford2021learning,alayrac2022flamingo}, to ensure that the resulting multi-modal representation encapsulates both the intricacies of language and the details of visual content:

\begin{equation}
\small
U_{mm} = \sum_{i=1}^{N} \text{softmax}(\text{emb}_\text{vis} \cdot \text{emb}_\text{text}^T) \cdot \text{emb}_\text{text}(i)
\end{equation}
where \(\text{emb}_\text{text}(i)\) and \(\text{emb}_\text{vis}(i)\) represent the embedding of the textual instruction and Queries for the \(i\)-th element respectively. Simultaneously, \(\text{emb}_\text{text}(i)\) serves as both the key (K) and value (V) in traditional attention mechanism, while \(\text{emb}_\text{vis}(i)\) functions as the query (Q). The textual instruction, after undergoing CMAA, transforms into $U_{mm}$. Subsequently, $U_{mm}$ concatenate onto the output of Queries in the form of Figure \ref{comparison.png}, culminating in the final integration of visual and textual elements. Detailed information about CMAA can be referred to in Appendix \ref{cmaa}.

\begin{figure*}[htp]
\centering
\includegraphics[width=1\textwidth]{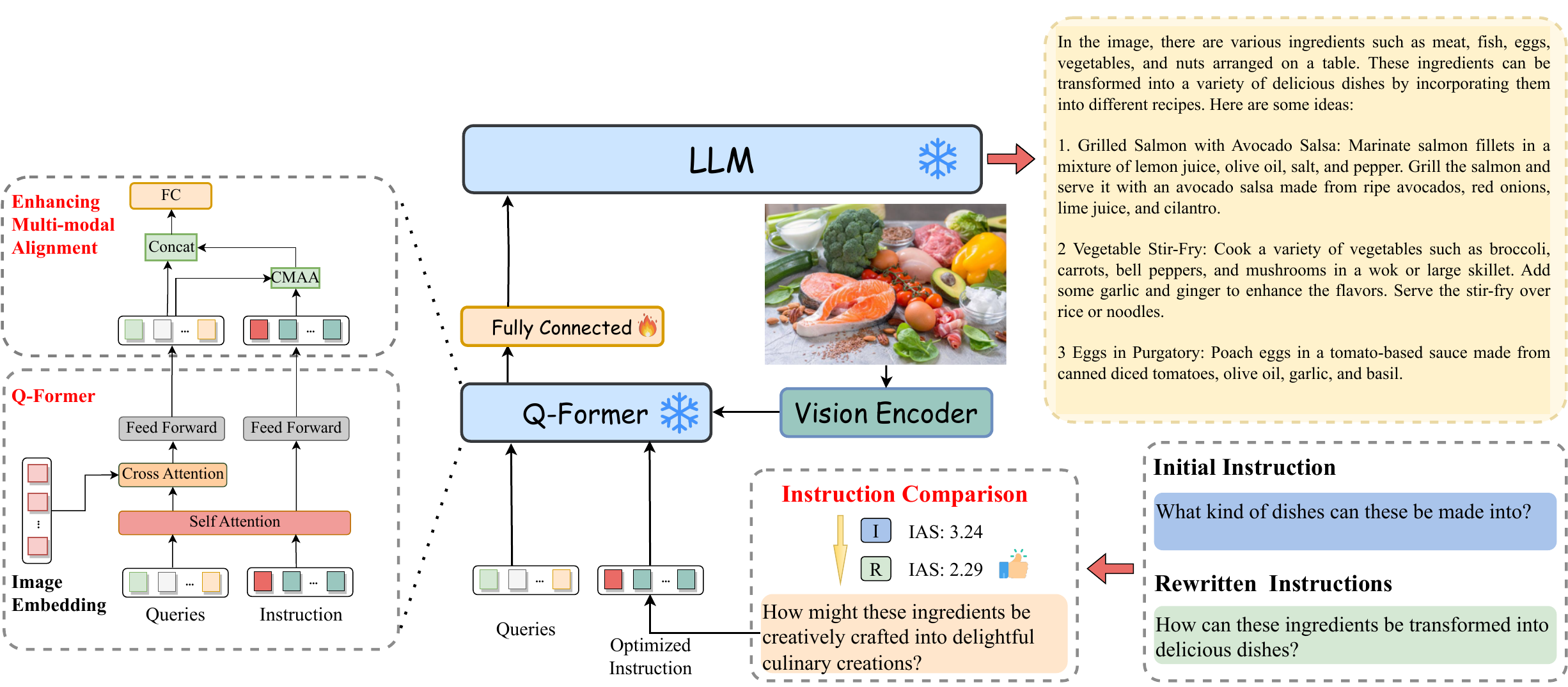}
\caption{The figure depicts the complete pipeline of Instruction Comparison Optimization. The initial and rewritten instructions are processed through comparison optimization to generate optimized instruction. Subsequently, the optimized instruction is utilized for generation in MMLMs.}
\label{optimization.png}
\end{figure*}

\textbf{Optimized Model Training and Performance:}
In developing the new architecture, our approach extends beyond mere technical integration to encompass optimization of training and performance. We employ selective weight freezing strategy, where specific layers of the pre-trained model are kept static to preserve learned features, and targeted fine-tuning, where newly introduced components or layers are specifically trained to adapt to the task at hand. This targeted approach allows us to fine-tune the model's performance without the need for extensive retraining \cite{hu2021lora}, thereby enhancing the learning efficiency and ensuring the robustness and scalability of the model \cite{toneva2018empirical,zhai2023investigating}. The objective function for training takes the following form:

\begin{equation}
p(\text{Y}_{text}|\text{X}_{img}) = \prod_{i=1}^{L}p_\theta(y_i|\text{X}_{img},\text{Y}_{text}^{[1:i-1]})
\end{equation}
where $\theta$ is the trainable parameters, $\text{X}_{img}$ and $\text{Y}_{text}$ respectively denote the input image and the output text, $\text{Y}_{text}^{[1:i-1]}$ represents the input instruction and the text already generated up to the $i-1$ step.

\subsection{Autonomous Instruction Optimization}
\label{sec:self_optimization}

During inference, the textual instruction has a significant impact on the generation results of MMLM. Therefore, we propose an approach that leverages the inherent text processing capabilities of MMLM to self-optimize textual instructions, thereby aligning the results more closely with user requirements. Our method comprises two stages: \textbf{Rewriting} Textual Instructions and Instruction \textbf{Comparison} Optimization.

\textbf{Rewriting Textual Instruction:} 
LLMs exhibit powerful text rewriting capabilities, preserving semantic information while modifying the content of the text. Therefore, our objective is to use the LLM in the MMLM to rewrite the initial textual instruction. The aim is obtain a pair of instructions that exhibit roughly equivalent semantics, thereby establishing a solid foundation for the next stage. It is important to note that the rewritten instruction that emerges from this process is not necessarily expected to surpass the initial instruction in quality. The mere occurrence of a difference between the pair is sufficient to satisfy the requirements of subsequent processes. This setting simplifies the text rewriting task, thereby lowering the barrier to its implementation.

Specifically, we designed a prompt tailored for LLM to rewrite the initial instruction. `Initial instruction' refers to the original instruction sent by the user. The prompt directs LLM on how to rewrite the initial instruction while ensuring minimal semantic changes between the initial and the rewritten versions. The template of the prompt used in this stage can be referred to in the Appendix \ref{app:instruction rewritingtemplates}. Notably, since this stage solely involves instruction rewriting, it does not necessitate the entire MMLM. Employing only the LLM part could marginally decrease the time consumed by the rewriting process.

\textbf{Instruction Comparison Optimization: } 
At this stage, we devise a method that allows the MMLM to identify the superior instruction via comparative analysis, with the aim to generate higher-quality instruction. As depicted in Figure \ref{icl.png}, we innovatively apply ICL to rank cases, enabling the model to ascertain the quality of instructions solely through comparison between the pair of initial and rewritten instructions \cite{ren2023context}. 

Considering that the ultimate purpose of the instructions is to aid inference by MMLM, we posit that the quality of these instructions should be evaluated by the MMLM themselves. Specifically, we enable MMLM to score the instruction independently, without the assistance of an external discriminator. As such, we proposed the Instruction Alignment Score (IAS), devised to measure the expected confidence of the evaluation instruction under the condition of a given image. We employ a prompt to guide MMLM in scoring the instruction. The template for this prompt can be found in Appendix \ref{app:MPG}. Defined as the expectation of negative log-probability, IAS is calculated as follows:

\begin{equation}
\small
    \mathrm{\mathrm{IAS}} = \mathbb{E}[-\text{log} P(t_i|\text{X}_{img},\text{X}_{prompt},t_{[1:i-1]};\theta)]
\label{eq3}
\end{equation}

Here, $\text{X}_{img}$ is the input image, $\text{X}_{prompt}$ denotes the prompt employed to guide the model in its computations, $\theta$ symbolizes our MMLM model and $t_i$ represents the tokens from the textual instruction the MMLM are evaluating for quality. The negative log-probability, which originally served as the loss function for LLMs, is utilized in Equation \ref{eq3} to assess the fluency of the given image and instruction under the current MMLM. A lower IAS indicates a higher alignment of the instruction with the model’s understanding, enabling MMLM to perform better. After calculating IAS, as shown in Figure \ref{icl.png}, we rank the \textbf{two} instruction-IAS pairs in descending order, and combine them into a prompt in the form of ICL. This is then input into MMLM to generate an optimized instruction. The optimized instruction will have better inference performance compared to the initial and rewritten instructions. The complete pipeline is presented in Figure \ref{optimization.png} and Appendix \ref{aio}.

\begin{table*}[htp]
    \small 
    \setlength{\tabcolsep}{4pt} 
    \renewcommand{\arraystretch}{1.2} 
    \centering
    \begin{tabularx}{\textwidth}{lcccccccccc}
        \toprule
        \multirow{2}{*}{} & \multicolumn{2}{c}{Image Captioning} & \multicolumn{3}{c}{Visual Reasoning} & \multicolumn{2}{c}{Image QA} & \multicolumn{3}{c}{Comprehensive VQA} \\
        \cmidrule(lr){2-3} \cmidrule(lr){4-6} \cmidrule(lr){7-8} \cmidrule(lr){9-11}
         & Flickr30K & Nocaps & VSR & GQA & IconQA & VizWiz & TextVQA & Visdial & SciQA & HM \\
        \midrule
       BLIP-2 ($\mathrm{FlanT5_{XXL}}$) & 73.7 & 104.5 & 68.2 & 44.6 & 45.4 & 29.4 & 44.1 & 46.9 & 64.5 & 52.0 \\
       BLIP-2 ($\mathrm{Vicuna_{13B}}$) & 74.9 & 107.5 & 50.9 & 41.0 & 40.6 & 19.6 & 42.5 & 45.1 & 61.0 & 53.7 \\
       MiniGPT-4 ($\mathrm{Vicuna_{13B}}$) & / & / & 50.7 & 30.8 & 37.6 & 34.8 & 18.7 & / & / & 29.0 \\
       LLaVA ($\mathrm{Vicuna_{13B}}$) & / & / & 56.3 & 41.3 & 43.0 & 37.7 & 28.3 & / & / & 9.2 \\
       InstructBLIP ($\mathrm{FlanT5_{XL}}$) & 84.5 & 119.9 & 64.8 & 48.4 & 50.0 & 32.7 & 46.6 & 46.6 & 70.4 & 56.6 \\
       InstructBLIP ($\mathrm{FlanT5_{XXL}}$) & 83.5 & 120.0 & 65.6 & 47.9 & 51.2 & 30.9 & 46.6 & 48.5 & 70.6 & 54.1 \\
       InstructBLIP ($\mathrm{Vicuna_{7B}}$) & 82.4 & 123.1 & 54.3 & 49.2 & 43.1 & 34.5 & 50.1 & 45.2 & 60.5 & 59.6 \\
       InstructBLIP ($\mathrm{Vicuna_{13B}}$) & 82.8 & 121.9 & 52.1 & 49.5 & 44.8 & 33.4 & 50.7 & 45.4 & 63.1 & 57.5 \\
       BLIVA ($\mathrm{Vicuna_{13B}}$) & 87.1 & / & 62.2 & / & 44.9 & 42.9 & 58.0 & 45.6 & / & 55.6 \\
       BLIVA ($\mathrm{FlanT5_{XXL}}$) & 87.7 & / & \textbf{68.8} & / & \textbf{52.4} & 44.0 & 57.2 & 36.2 & / & 50.0 \\
       \midrule
       Ours($\mathrm{FlanT5_{XL}}$)  & 85.3 & 119.5 & 64.1 & 47.9 & 50.4 & 33.0 & 48.7 & 47.0 & 71.0 & 60.0 \\
       Ours($\mathrm{FlanT5_{XXL}}$)  & \textbf{88.5} & 120.4 & 66.9 & 48.1 & 51.2 & 31.3 & 48.8 & \textbf{49.2} & \textbf{81.8} & 55.7 \\
       Ours($\mathrm{Vicuna_{7B}}$)  & 87.9 & \textbf{124.2} & 60.1 & 52.0 & 44.2 & 42.7 & 60.6 & 45.7 & 74.6 & \textbf{62.7} \\
       Ours($\mathrm{Vicuna_{13B}}$)  & 84.0 & 119.8 & 56.2 & \textbf{52.9} & 50.3 & \textbf{45.0} & \textbf{65.6} & 45.7 & 71.0 & 58.9 \\
       \bottomrule
    \end{tabularx}
    \caption{Zero-shot results on general image-text benchmarks. Here, Visdial, SciQA, and HM respectively refer to Visual Dialog, ScienceQA, and HatefulMemes. The results for MiniGPT-4 and LLaVA are sourced from BLIVA \cite{hu2023bliva}, while the remaining results originate from their respective papers \cite{li2023blip,instructblip}.}
    \label{tab:zero-shot}
\end{table*}

\section{Experiments}

\subsection{Datasets}

The datasets in this paper primarily consists of a training dataset and the zero-shot evaluation benchmarks. The training data is sourced from LLaVA, which is also a subset of the InstructBLIP training datasets. The data was collected by the authors of LLaVA using ChatGPT/GPT-4 \cite{ChatGPT,GPT-4}, following a multi-modal instruction format. We believe that using the same dataset as previous work enables a fairer comparison in the experiments. In Appendix \ref{app:training datasets}, this paper provides more details related to the training dataset.

For zero-shot evaluation benchmarks, to ensure alignment for comparison, we also follow InstructBLIP. The evaluation domains include: Image captioning: Flickr30K \cite{young2014image}, NoCaps \cite{agrawal2019NoCaps}. Visual Reasoning: VSR \cite{liu2023vsr}, GQA \cite{hudson2019gqa}, IconQA \cite{lu2021iconqa}. Image QA: VizWiz \cite{gurari2018vizwiz}, TextVQA \cite{mishra2019ocr}. Comprehensive VQA: Visual Dialog \cite{das2017visual}, ScienceQA \cite{lu2022learn}, HatefulMemes \cite{kiela2020hateful}. In the Appendix \ref{app:datasets detailss}, we provide the details of the evaluation benchmarks as comprehensively as possible.

\subsection{Implementation Details}

In terms of the model architecture, we opted for the ViT-G/14 from EVA-CLIP \cite{fang2023eva} as the visual encoder, removing the final layer of the ViT and utilizing the output features from the penultimate layer. In line with InstructBLIP, we employed two distinct LLMs: FlanT5 and Vicuna. FlanT5, derived from the instruction-tuning of the encoder-decoder Transformer T5 \cite{raffel2020exploring}, encompasses two sizes: FlanT5-XL and FlanT5-XXL. Vicuna, on the other hand, is refined from the instruction-tuning of the decoder-only Transformer LLaMA \cite{touvron2023llama}, and also includes two sizes: Vicuna-7B and Vicuna-13B. The weights of both Q-Former and the fully connected layers are sourced from InstructBLIP and need to correspond to different LLMs. Our entire model framework requires freezing the weights of the visual encoder, Q-Former, and LLMs, allowing only the fully connected layers to be unfrozen. Further details regarding training hyperparameters can be found in Appendix \ref{app:training details}.

\subsection{Zero-shot Evaluation}

We conducted zero-shot learning of our model against previous state-of-the-art (SOTA) works across 10 benchmarks in Table \ref{tab:zero-shot}. It's evident that our model showcases a significant advantage in the majority of benchmarks, especially in Image QA and Comprehensive VQA domains. Specifically, our methods has improved the previous SOTA results by 13.1\% and 9\% in TextVQA and HatefulMemes. Furthermore, as our model weights are primarily inherited from InstructBLIP, a side-by-side comparison with InstructBLIP indicates that our method significantly enhances the overall capability of MMLMs. For example, based on the FlanT5-XXL model, our method improved upon InstructBLIP by 6\% and 15.9\% on Flickr30K and ScienceQA, respectively.

The results in Table \ref{tab:zero-shot} indicate that our proposed instruction optimization method exhibits very significant gains for image-text tasks. However, a small portion of the evaluation results shows some discrepancies with the overall trend. Specifically, our method exhibits very limited or even inferior performance compared to the baseline on NoCaps dataset. We attribute this phenomenon to potential biases introduced by the training set. The training set of InstructBLIP is much richer than ours, and fine-tuning solely on its subset may lead to a certain degree of catastrophic forgetting. Furthermore, another issue arises in the HatefulMemes where the smaller LLM backbone works better. The primary reason for this phenomenon might be the insufficient magnitude of parameter difference between LLMs, failing to establish a clear dominance. This observation is similarly reflected in the performance on InstructBLIP. 

\begin{figure*}[htp]
\centering
\includegraphics[width=1\textwidth]{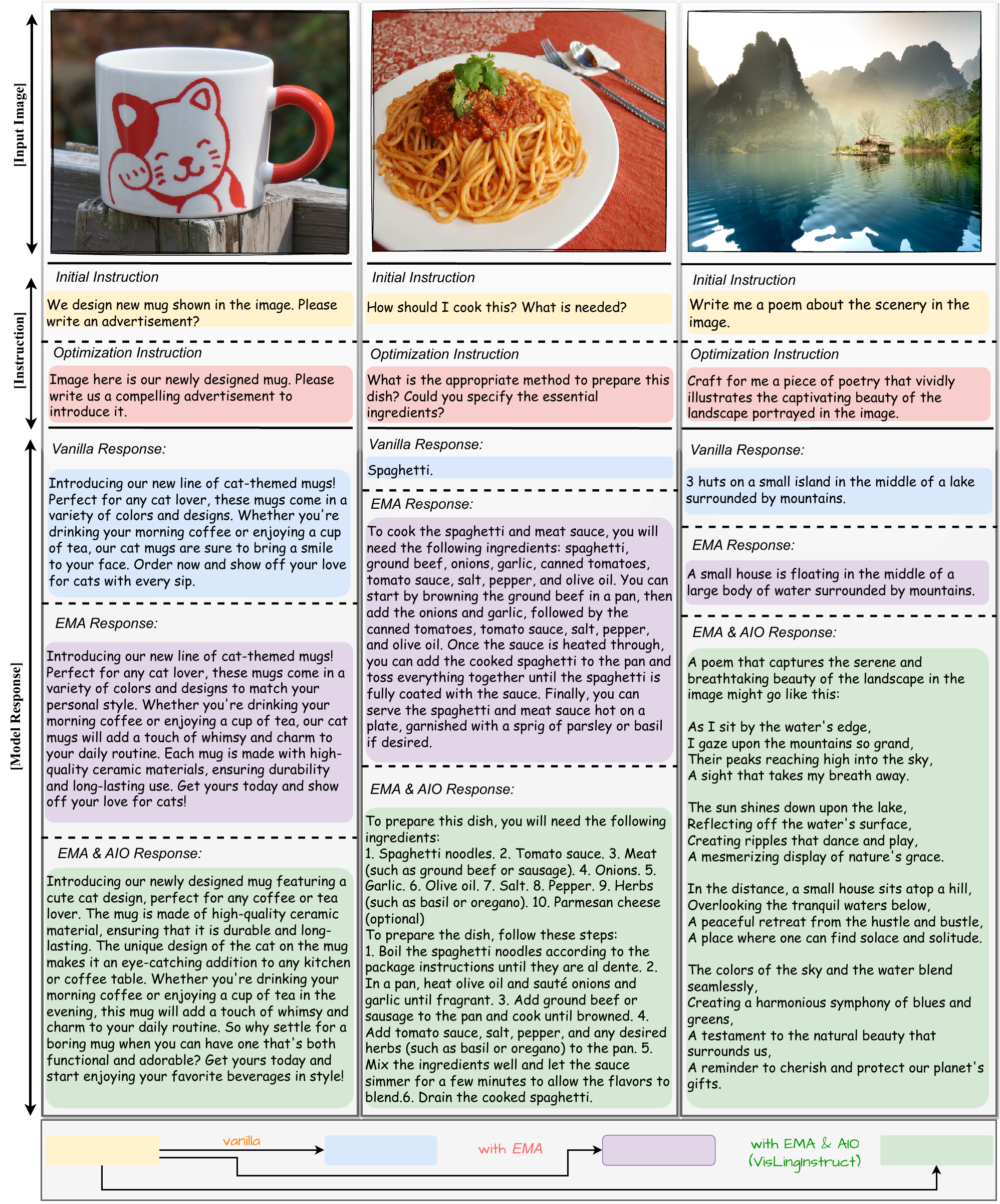}
\caption{The one on the left is a case written for a product advertisement, the one in the middle is a recipe description, and the one on the right is a poetry creation. Qualitative comparison of three responses from different ablations: initial instruction with vanilla model (blue), initial instruction with EMA model (purple), and optimized instruction with EMA model (green).}
\label{img:qualitative}
\end{figure*}

\begin{table*}[t]
    \small 
    \setlength{\tabcolsep}{4pt} 
    \renewcommand{\arraystretch}{1.2} 
    \centering
    \resizebox{\textwidth}{!}{
    \begin{tabular}{c c c c|cccccccccc}
       \toprule
       \multicolumn{1}{c}{\textbf{Vanilla}} & \multicolumn{1}{c}{\textbf{EMA}} & \multicolumn{2}{c|}{\textbf{AIO}} & \multicolumn{2}{c}{Image Captioning} & \multicolumn{3}{c}{Visual Reasoning} & \multicolumn{2}{c}{Image QA} & \multicolumn{3}{c}{Comprehensive VQA} \\ \cmidrule(lr){3-4} \cmidrule(lr){5-6} \cmidrule(lr){7-9} \cmidrule(lr){10-11} \cmidrule(lr){12-14}
       \multicolumn{1}{c}{} & \multicolumn{1}{c}{} & \multicolumn{1}{c}{Rewriting} & \multicolumn{1}{c|}{Comparison} & Flickr30K & NoCaps & VSR & GQA & IconQA & VizWiz & TextVQA & Visdial & SciQA & HM \\
        \midrule
        \multicolumn{12}{c}{\textbf{\textit{FlanT5-XL}}} \\
       \Checkmark & & & & 84.5 & \textbf{119.9} & 64.8 & 48.4 & 50.0 & 32.7 & 46.6 & 46.6 & 70.4 & 56.6 \\
       \Checkmark & \Checkmark & & & 85.1 & 119.7 & 63.5 & \textbf{48.6} & 50.0 & 32.8 & 48.5 & 46.9 & 70.6 & \textbf{60.8} \\
       \Checkmark & \Checkmark & \Checkmark & & 84.7 & 118.1 & \textbf{66.8} & 48.5 & 49.0 & 31.8 & 47.5 & 44.8 & 70.4 & 57.3 \\
       \Checkmark & \Checkmark & \Checkmark & \Checkmark & \textbf{85.3} & 119.5 & 64.1 & 47.9 & \textbf{50.4} & \textbf{33.}0 & \textbf{48.7} & \textbf{47.0} & \textbf{71.0} & 60.0 \\
         \midrule
         \multicolumn{12}{c}{\textbf{\textit{FlanT5-XXL}}} \\
       \Checkmark & & & & 83.5 & 120.0 & 65.6 & 47.9 & 51.2 & 30.9 & 46.6 & 48.5 & 70.6 & 54.1 \\
       \Checkmark & \Checkmark & & & 86.3 & 120.3 & 55.7 & 48.0 & \textbf{51.6} & \textbf{31.5} & 48.3 & 49.0 & \textbf{82.0} & 55.2 \\
       \Checkmark & \Checkmark & \Checkmark & & 85.3 & 120.1 & 66.5 & 48.1 & 50.9 & 31.1 & 46.7 & 48.5 & 73.5 & 54.1 \\
       \Checkmark & \Checkmark & \Checkmark & \Checkmark & \textbf{88.5} & \textbf{120.4} & \textbf{66.9} & \textbf{48.3} & 51.2 & 31.3 & \textbf{48.8} & \textbf{49.2} & 81.8 & \textbf{55.7} \\
         \midrule
         \multicolumn{12}{c}{\textbf{\textit{Vicuna-7B}}} \\
       \Checkmark & & & & 82.4 & 123.1 & 54.3 & 49.2 & 43.1 & 34.5 & 50.1 & 45.2 & 60.5 & 59.6 \\
       \Checkmark & \Checkmark & & & 81.6 & 124.5 & \textbf{60.6} & 51.9 & 43.2 & 40.5 & 49.9 & 45.3 & 55.4 & 60.8 \\
       \Checkmark & \Checkmark & \Checkmark & & 82.3 & \textbf{124.5} & 55.4 & 47.6 & 44.0 & 40.3 & 58.3 & 43.4 & 63.0 & 62.2 \\
       \Checkmark & \Checkmark & \Checkmark & \Checkmark & \textbf{87.9} & 124.2 & 60.1 & \textbf{52.0} & \textbf{44.2} & \textbf{42.7} & \textbf{60.6} & \textbf{45.7} & \textbf{74.6} & \textbf{62.7} \\
         \midrule
         \multicolumn{12}{c}{\textbf{\textit{Vicuna-13B}}} \\
       \Checkmark & & & & 82.8 & \textbf{121.9} & 52.1 & 49.5 & 44.8 & 33.4 & 50.7 & 45.4 & 63.1 & 57.5 \\
       \Checkmark & \Checkmark & & & 84.4 & 120.2 & \textbf{58.9} & 51.6 & 48.4 & 43.0 & 56.9 & 43.0 & 48.4 & \textbf{61.0} \\ 
       \Checkmark & \Checkmark & \Checkmark & & 80.4 & 120.6 & 52.5 & 51.1 & 49.3 & 41.5 & 62.4 & 44.4 & 68.0 & 58.7 \\ 
       \Checkmark & \Checkmark & \Checkmark & \Checkmark & \textbf{84.0} & 120.8 & 56.2 & \textbf{52.9} & \textbf{50.3} & \textbf{45.0} & \textbf{65.6} & \textbf{45.7} & \textbf{71.0} & 58.9 \\
        \bottomrule
    \end{tabular}
    }
    \caption{Results of ablation studies for Enhancing Multi-modal Alignment (EMA) and Autonomous Instruction Optimization (AIO) in different LLM backbones. Among them, EMA is split into Rewriting Textual Instructions (Rewriting) and Instruction Comparison Optimization (Comparison) for discussion respectively. Vanilla represents the baseline model without any of our proposed modules and \Checkmark indicates that the module has been integrated.}
    \label{tab:ablation_big}
\end{table*}

\subsection{Ablation Study}

To investigate the impact of EMA (Section \ref{sec:architecture}) and AIO (Section \ref{sec:self_optimization}) on the final results, we conducted ablation studies by individually removing them during evaluation. 

As depicted in Table \ref{tab:ablation_big}, after integrating the EMA mechanism on the vanilla baseline, the overall performance of all models is significantly enhanced. This indicates that our EMA method indeed enhances the alignment between images and text. Moreover, if AIO continues to be integrated on the basis of EMA, the evaluation results can be further improved. This adequately shows that the two mechanisms can strengthen each other. EMA, by enhancing its perception of instructions, can serve as a booster to further enhance AIO. 

As for the AIO part, we also further split it to conduct ablation experiments. We discuss Rewriting Textual Instructions and Instruction Comparison Optimization separately. It can be clearly seen from the results in Table \ref{tab:ablation_big} that instruction rewriting cannot continue to improve the effect on the basis of EMA. On the contrary, it is even inferior to the vanilla baseline in many results. This phenomenon fully demonstrates that just rewriting cannot stably optimize the instruction, and requires correction by our Instruction Comparison Optimization mechanism.

Additionally, a particular phenomenon is observed in Table \ref{tab:ablation_big}, where the encoder-decoder FlanT5 and the decoder-only Vicuna exhibit slight inconsistencies when our methods are applied. For instance, EMA is beneficial on the ScienceQA dataset for FlanT5, but performs poorly for Vicuna. The reasons for this phenomenon might be firstly because LLMs with different structures excel at different tasks. From our general understanding of model structures, the encoder part from FlanT5 is more suitable for tasks involving feature comprehension. Secondly, the corpus used during the pre-training of the model is also a crucial factor. FlanT5 might perform better in certain tasks simply because the model has encountered related content during pre-training. 

We also conducted experiments and analyses on the number of instructions in ICL and the computational overhead of the proposed method. Detailed reports of these studies can be found in Appendices \ref{sec:number} and \ref{sec:computational}, respectively.

\subsection{Qualitative Evaluation}

Beyond the benchmarks-driven experimental analyses, we diversified our qualitative evaluation by incorporating real-world images and instructions. As shown in Figure \ref{img:qualitative}, we have enumerated three cases for comprehensive analysis. The process commences with the input of an image, subsequent questions and answers revolve around this visual context. This is followed by the presentation of instructions, encompassing both the initial instructions and the optimized by the AIO module. Conclusively, model response is delineated. The output section for evaluation includes:  the results obtained by inputting the initial instructions into the vanilla model (Vanilla Response); the results obtained by inputting the initial instructions into the integrated EMA module model (EMA Response); and the results from inputting the optimized instructions into the integrated EMA module model (EMA \& AIO  Response), which is VisLingInstruct.

The outcome as observed in the figure suggests that the EMA Response demonstrates an improvement over the Vanilla Response, both in terms of content accuracy and richness of detail. For instance, within the case of poetry creation, the erroneously presented ‘3 huts’ is accurately identified as ‘a small house’. In the case of recipe description, the narrative about spaghetti is much more detailed in the EMA Response. Furthermore, the EMA \& AIO response also surpasses the EMA response alone, evident in the former's answers possessing a superior logical organization and better fulfillment of user intent. This is well illustrated in all three cases presented in the figure. And for more on the performance in multi-turn dialogues, we have provided a demonstration and discussion in the Appendix \ref{app:dialogues}.

\section{Conclusion}

This paper proposes VisLingInstruct, a novel autonomous instruction optimization framework for visual-linguistic multi-modal models. We conducted a comprehensive study on multi-modal models and demonstrated the powerful autonomous instruction optimization capabilities of the VisLingInstruct model, demonstrating strong zero-shot learning capabilities in a series of benchmarks. At the end of the experiment, qualitative examples were used to demonstrate the specific situation of VisLingInstruct in autonomous instruction optimization, such as knowledge-based image description, image-based text creation and multi-turn dialogue. We hope that VisLingInstruct can inspire more new research on autonomous optimization of multi-modal instruction.

\section*{Limitations}

Despite VisLingInstruct is an effective method for MMLMs, it still possesses certain limitations, which include:

Firstly, our autonomous instruction optimization framework has a relatively large computational overhead. We have carried out a comprehensive discussion on this subject in Appendix \ref{sec:computational}. While we maintain that the added computations are quite justifiable and beneficial, it is undeniable that they augment the time required to yield MMLM results. We propose that future research should focus on optimizing the process of instruction optimization. We believe that such advancements will undoubtedly enhance the applicability and promotion of this technology.

Secondly, the experimental work presented within this paper is primarily concentrated on image and text modalities. There is also a real need to optimize instructions for other modalities. Consequently, we have earmarked this as future work, with the objective of verifying the efficacy of our framework on additional modalities, including video and audio.

\bibliography{anthology,custom}
\clearpage

\appendix
\section{Algorithm}

The algorithmic core of our approach in VisLingInstruct is structured around two main processes: Cross-Modal Alignment Attention and Autonomous Instruction Optimization. The former process harmonizes the integration of text and image, while the latter refines the textual instructions for MMLMs.

\subsection{Cross-Modal Alignment Attention}\label{cmaa}
The Cross-Modal Alignment Attention (CMAA) algorithm focuses on the integration of textual and visual embeddings, creating a unified text representation. 

\begin{algorithm}
\caption{Cross-Modal Alignment Attention}
\begin{algorithmic}[1]
\Require Textual embeddings $E_{\text{text}}$, Queries embeddings $E_{\text{que}}$
\Ensure Unified multi-modal representation $U_{\text{mm}}$
\State Initialize cross-modal alignment mechanism
\For{each element $i$ in $E_{\text{text}}$}
    \State Compute attention between $E_{\text{text}}(i)$ and $E_{\text{que}}$
    \State Assign attention weight on $E_{\text{text}}(i)$
\EndFor
\State $U_{\text{mm}} \leftarrow$ Aggregate of aligned and weighted $E_{\text{text}}$
\Return $U_{\text{mm}}$
\end{algorithmic}
\end{algorithm}


\subsection{Autonomous Instruction Optimization}\label{aio}
The Autonomous Instruction Optimization (AIO) is designed to transform initial instruction into an optimized format. 

\begin{algorithm}
\caption{Autonomous Instruction Optimization}
\begin{algorithmic}[1]
\Require Initial instructions $I_{i}$
\Ensure optimized instruction $I_{opt}$
\State Initialize autonomous instruction optimization
\State Rewriting the initial instruction $I_{i}$ to obtain $I_{j}$
\State Calculating the IAS for $I_{i}$ and $I_{j}$
\State Ranking the instruction-IAS pairs
\State $I_{refined} \leftarrow$ Constructing the prompt input for Instruction Comparison in MMLMs
\Return $I_{refined}$
\end{algorithmic}
\end{algorithm}

\section{Templates}

\subsection{Instruction Rewriting Templates}
\label{app:instruction rewritingtemplates}

Here is the template used for Instruction rewriting in this paper, where ‘\{\}’ signifies the instruction that requires modification:

\texttt{There is the text \{\}. Please modify the text to make it better while retaining the sentence structure and keywords.}

\subsection{IAS templates}
\label{app:MPG}

In the following prompt template, \{\} is used to place instructions requiring MPG calculation.

\texttt{<Image>Based on the image given, the most appropriate instruction should be: \{\}}

\section{Data and Training Details}

\subsection{Training Dataset Format}\label{app:training datasets}

For an image $X_v$, there is an associated question-answer pair <$X_q$, $X_a$> related to $X_v$. In some cases, there are multi-turn dialogues represented as (<$X_q^1$, $X_a^1$>,...,<$X_q^m$, $X_a^m$>). During training, for single-turn dialogue data, $X_q$ serves as the input instruction, while $X_a$ corresponds to the ground truth. Likewise, for multi-turn dialogue data, it is essential to concatenate the historical dialogues (excluding the last turn) and append them along with $X_q^m$ as the input. Meanwhile, $X_a^m$ serves as the ground truth.

\subsection{Zero-shot Evaluation Datasets Details}
\label{app:datasets detailss}

As shown in Table \ref{tab:benchmarks}, the evaluation parts chosen by different benchmarks are not the same. We have adopted the settings from InstructBLIP. It's important to note that for ScienceQA, we only evaluate the set with image context. The evaluation metrics vary across benchmarks: NoCaps and Flickr30K employ CIDEr scores \cite{vedantam2015cider}, HatefulMemes utilizes AUC scores, and Visual Dialog employs Mean Reciprocal Rank (MRR). For all remaining datasets, top-1 accuracy is used as the metric. All evaluation benchmarks have no data overlap with the training set, ensuring the authenticity of zero-shot. 

Table \ref{tab:optimization} illustrates the initial instructions for all benchmarks. The initial instructions were predominantly sourced from InstructBLIP. ‘\{\}’ contains entities such as questions from each individual case. For instructions with options, we separate the choices alphabetically, for instance: (a) apple (b) banana (c) pineapple.

\begin{table}[htp]
    \centering
    \begin{tabular}{lcc}
        \toprule
        \textbf{Dataset Name}  & \textbf{Part} & \textbf{count} \\
        \midrule
        Flickr30K & test & 1000 \\
        NoCaps & val & 4500 \\
        VSR & test & 1222 \\
        GQA & test-dev & 12578 \\
        IconQA & test & 6316 \\
        VizWiz & test-dev & 4319 \\
        TextVQA & val & 5000 \\
        Visual Dialog  & val & 2064 \\
        ScienceQA & test & 2017 \\
        HatefulMemes & val & 1040 \\
        \bottomrule
    \end{tabular}
    \caption{The selected part in all zero-shot evaluation benchmarks, and accompanied by specific data count.}
    \label{tab:benchmarks}
\end{table}

\begin{table*}[t]
    \centering
    \begin{tabularx}{\textwidth}{p{0.1\textwidth}|p{0.7\textwidth}}
        \toprule
        \textbf{Dataset}  & \textbf{Initial instruction} \\
        \hline
        Flickr30K/\newline NoCaps & <Image>A short image description: \\
        \hline
        VSR & <Image>Based on the image, is this statement true or false? \{\} \\
        \hline
        GQA/\newline Visdial & <Image>Question: \{\} \textbackslash n Short answer: \\
        \hline
        IconQA & <Image>Question: \{\} Options: \{\} \textbackslash n Answer: \\
        \hline
        VizWiz & <Image>Answer the question based on the image. Reply in one phrase/word or say ‘unanswerable’. Question: \{\} \textbackslash n Short answer: \\
        \hline
        TextVQA & <Image>OCR tokens: \{\} Question: \textbackslash n Short answer: \\
        \hline
        SciQA & <Image>Given the image, choose the correct option for the following question. Question: \{\} \textbackslash n Options: \{\} \\
        \hline
        HM & <Image>This is an image with: \{\} written on it. Is it hateful? \\
        \bottomrule
    \end{tabularx}
    \caption{Presentation of initial instructions for each benchmark.}
    \label{tab:optimization}
\end{table*}

\subsection{Training Details}
\label{app:training details}

We implement VisLingInstruct by LAVIS library \cite{li2022lavis}. We fine-tuned the fully connected layers for 3 epochs, employing different hyperparameters across distinct LLMs. We employ a batch size of 32, 128 and 256 for the Vicuna-7B/13B, FlanT5-XL and FlanT5-XXL, respectively. For each model, we conduct validation every 1K steps. Our training procedures was the utilization of the AdamW \cite{loshchilov2018decoupled} optimizer with $\beta_1$ = 0.9, $\beta_2$ = 0.999, and a weight decay of 0.05. We implemented a linear warm-up of the learning rate over the initial 1K steps, escalating from $10^-8$ to $10^-5$, followed by cosine decay towards a minimum learning rate of 0. All our model's trainable parameter counts are maintained within the range of a few million. Under the conditions of 8 A100 40G, the training durations for FlanT5, Vicuna 7B, and Vicuna 13B are 105 minutes, 135 minutes, and 210 minutes.

During the evaluation process, we employed two different generation methods tailored to different benchmarks. For the domain of benchmarks such as Image Captioning, results were directly generated from instructions. These results were then compared against ground truth to calculate metrics. On the other hand, for classification-based VQA tasks, we followed previous work \cite{alayrac2022flamingo,instructblip} by computing the language model loss for each candidate option and selecting the one with the lowest loss as the final prediction. This method was applied to ScienceQA, IconQA, HatefulMemes, and Visual Dialog.

\begin{table}
    \small
    \centering
    \begin{tabularx}{\columnwidth}{lXXXX}
    \toprule
        Backbone & 1R & 2R & 3R & 4R \\
    \midrule
    \multicolumn{5}{c}{\textbf{\textit{Rewriting}}} \\
       FlanT5-XL  & 62.7 & 61.1 & 59.4 & 57.0 \\
       FlanT5-XXL  & 64.2 & 63.6 & 62.7 & 58.9 \\
       Vicuna-7B  & 65.5 & 64.8 & 63.2 & 60.4 \\
       Vicuna-13B  & 64.9 & 62.9 & 62.4 & 60.1 \\
    \midrule
    \multicolumn{5}{c}{\textbf{\textit{Loop}}} \\
       FlanT5-XL  & 62.7 & 61.5 & 60.2 & 58.4 \\
       FlanT5-XXL  & 64.2 & 64.4 & 62.6 & 59.5 \\
       Vicuna-7B  & 65.5 & 65.8 & 63.9 & 61.0 \\
       Vicuna-13B  & 64.9 & 63.7 & 62.1 & 60.3 \\
    \bottomrule
    \end{tabularx}
    \caption{The results about the impact of the number in the ICL instruction comparison with different LLM backbones. 1R represents that we do not add new instructions, which is the standard setting in our method. 2R to 4R represent the corresponding rounds of instruction generation. All the results are the average values of 10 benchmarks.}
    \label{tab:more_instructions}
\end{table}

\section{More Experiments and Analyses}

\subsection{Number of Instructions in ICL}\label{sec:number}

The process of instruction comparison is a crucial step, therefore we have explored the possibility of adding more instructions into ICL. In particular, we adopted two verification methods: firstly, we let the LLM generate multiple different \textbf{rewritten} instructions to increase the number of instructions involved in the ICL. Secondly, we continue to add optimized instructions generated by MMLM to the ICL for comparison and then generate new optimized instructions in a \textbf{loop}. 

However, as shown in Table \ref{tab:more_instructions}, neither of these operations could enhance the optimization effect of the instructions. On the contrary, the effect deteriorates as the number of rounds increases. We analyzed the possible reason is that the initial instruction is issued by the user, while the rewritten and optimized instructions are generated by MMLM. The statistical distribution of the user’s instruction is significantly different from those generated by MMLM. The larger the difference, the greater the benefit derived from the comparison. However, the distribution of instructions generated by MMLM is similar, introducing them into the ICL for mutual comparison may introduce undesirable noise. 

Meanwhile, the first method saw a more severe decline in effectiveness compared to the second. This is likely because the distributional difference between the additional instructions generated by rewriting is even smaller than that produced by looping, rendering the effect of ICL comparison almost negligible.

\begin{table}
    \small
    \centering
    \begin{tabularx}{\columnwidth}{lcccc}
    \toprule
       Backbone  & NoCaps & VSR & TextVQA & HM \\
    \midrule
       FlanT5-XL  & 2.4 & 6.7 & 4.5 & 7.0 \\
       FlanT5-XXL  & 2.8 & 6.6 & 4.2 & 7.3 \\
       Vicuna-7B  & 2.7 & 7.1 & 4.3 & 7.2 \\
       Vicuna-13B  & 2.6 & 6.9 & 4.6 & 7.4 \\
    \bottomrule
    \end{tabularx}
    \caption{The table records the computational overhead of VisLingInstruct on relevant benchmarks. We have selected four that are representative of their respective domains from ten benchmarks. The results denote that the multiple of time required by VisLingInstruct to complete the dataset compared to the time taken by the vanilla baseline.}
    \label{tab:computational}
\end{table}

\subsection{Computational Overhead}\label{sec:computational}

We conducted a comprehensive analysis of the computational overhead of our proposed method. VisLingInstruct, in the process of optimizing the initial instruction input by users, introduces some intermediate results, resulting in increased computation. For sample, the LLM from MMLM first generates the rewritten instruction based on the initial. Then, the MMLM needs to calculate the IAS of the two instructions. This step can rely on parallel computing, so it is equivalent to one computing time. Finally, the MMLM needs to further generate a refined instruction before finally producing a result. Therefore, the time cost is actually 3 times that of the vanilla baseline. 

The aforementioned is purely a theoretical presumption. To quantify the specific computational overhead, we conducted several experiments. As depicted in Table \ref{tab:computational}, the time overhead varies across different benchmarks. This variability can be attributed to the fact that the intermediate results of our proposed method are all instruction-related, and the length of the result returned by the model is the ultimate determinant of the computation time. For instance, in the VSR and HM tasks, the model only needs to respond with a simple `yes' or `no'. As this is significantly shorter than the input instruction, the additional computation associated with the instruction becomes markedly impactful. Conversely, in tasks like NoCaps, the result returned by the model surpasses the length of the input instruction. As a result, the overall computational overhead of VisLingInstruct is diluted to less than 3 times that of the vanilla baseline.

The engineering of prompt and instruction always introduces additional computational overhead, which is inevitable. The key question is whether such overhead is worthwhile. In the real-world demands of MMLMs, instructions are often much shorter compared to the final generated results, such as advertisements or copywriting based on product images. Compared to the time spent producing results, the time cost centered around the instruction is not unacceptable. On the contrary, users maybe get a better experience because they get better results.

\subsection{Multi-turn dialogues case studies}
\label{app:dialogues}

\begin{figure*}[htp]
\centering
\includegraphics[width=1\textwidth]{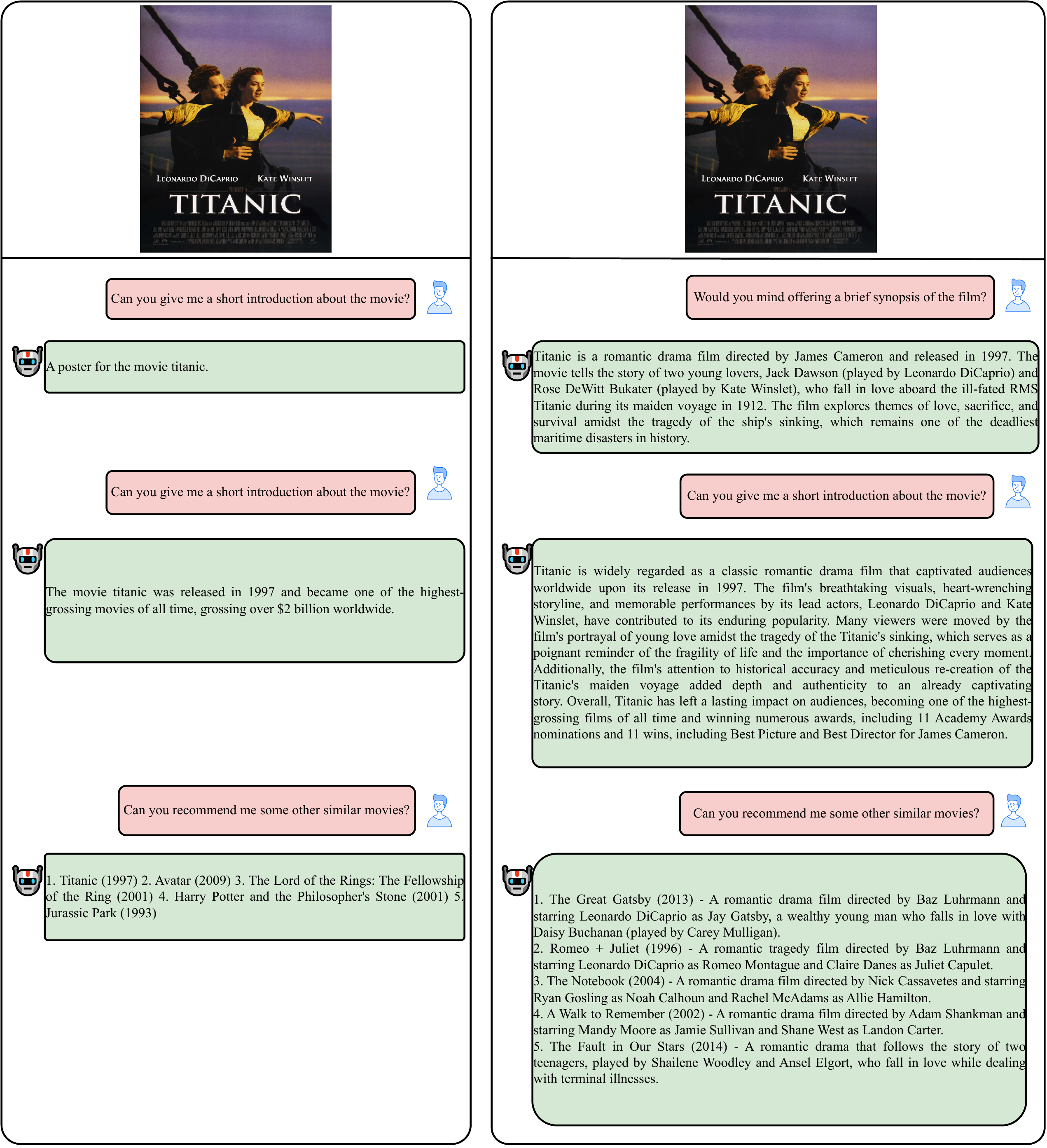}
\caption{This multi-turn dialogue example demonstrates the effectiveness of VisLingInstruct in movie introduction and recommendation. Our generated results (right) exhibit significantly improved accuracy in identifying instructional content and richness in answers compared to the vanilla model's results (left).}
\label{img:conversation1}
\end{figure*}

\begin{figure*}[htp]
\centering
\includegraphics[width=1\textwidth]{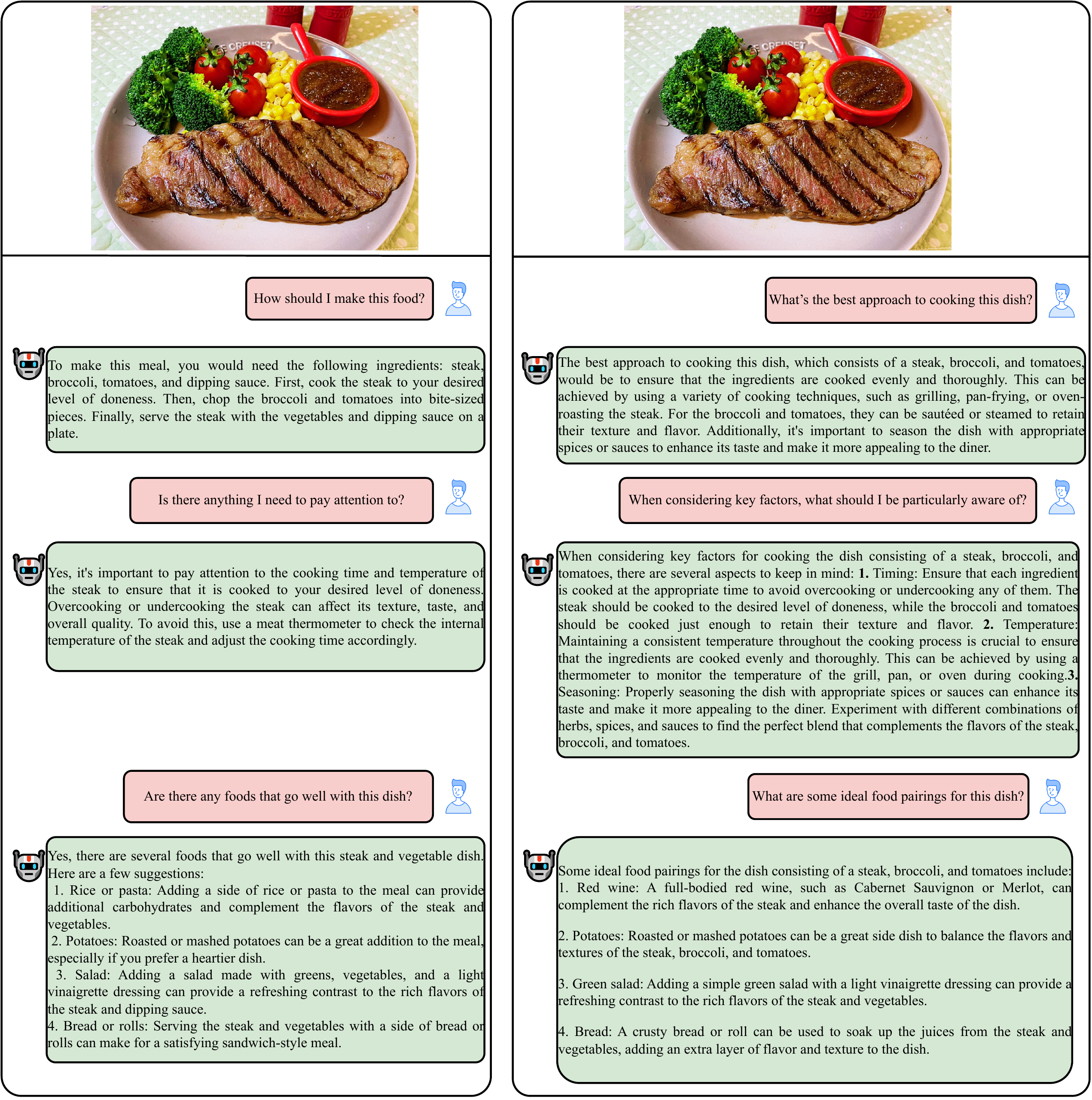}
\caption{This is a multi-turn dialogue about cooking topics. The two generated answers exhibit similar accuracy in content recognition. However, in terms of richness and the final food pairing, our response (right) surpasses the vanilla model's response (left) significantly.}
\label{img:conversation2}
\end{figure*}

As illustrated in Figures \ref{img:conversation1} and \ref{img:conversation2}, we present additional cases regarding multi-turn dialogue to showcase the strengths and distinctive features of our approach. The multi-turn generation results on the right originate from the models incorporating the EMA and AIO proposed in this paper (VisLingInstruct), while those on the left exclude this aspect, representing the results from Vanilla Response.

\end{document}